\pdfoutput=1
\documentclass[11pt]{article}

\usepackage{emnlp2021}

\usepackage{times}
\usepackage{latexsym}

\usepackage[T1]{fontenc}
\usepackage[utf8]{inputenc}

\usepackage{microtype}
\usepackage{graphicx}
\usepackage{bm}
\usepackage{amsmath}
\usepackage{amsfonts}
\usepackage{bbm}

\usepackage{helvet}
\usepackage{courier}
\usepackage{multirow}
\usepackage{bm}
\usepackage{amsmath}
\usepackage{mdwlist}

\usepackage{relsize}
\usepackage{amsfonts}
\usepackage{url}
\usepackage{graphicx}
\usepackage{subfigure}
\usepackage{caption}
\usepackage{hhline}
\usepackage{color}
\usepackage{colortbl}
\usepackage{booktabs}
\usepackage{tabularx}
\usepackage{amsmath}
\usepackage[linesnumbered,boxed,ruled,commentsnumbered]{algorithm2e}

\title{A Relation-Oriented Clustering Method for Open Relation Extraction}
\author{Jun Zhao$^{1}$,\ \ Tao Gui$^{2}$\footnotemark[1],\ \ Qi Zhang$^{1}$\footnotemark[1],\ \  Yaqian Zhou$^1$\\
  $^1$School of Computer Science, Shanghai Key Laboratory of Intelligent Information Processing,\\
  Fudan University, Shanghai, China\\
  $^2$Institute of Modern Languages and Linguistics, Fudan University\\
  \texttt{\{zhaoj19,tgui,qz,zhouyaqian\}@fudan.edu.cn}}

\begin{document}
\maketitle

\renewcommand{\thefootnote}{\fnsymbol{footnote}}
\footnotetext[1]{Corresponding authors.}
\begin{abstract}

%The clustering-based unsupervised relation discovery method has gradually become one of the important methods of open relation extraction (OpenRE). However, high-dimensional vectors can encode complex linguistic information, a mixture of which is leveraged by current unsupervised methods, which leads to the problem that the similarity between vectors does not accurately reflect the similarity of semantic relations between given entity pairs. In this work, we propose a relation-oriented clustering model and use it to identify the novel relations in the unlabeled data. To maximize the representation similarity of the instances with same relation, our method leverages the readily available labeled data of pre-defined relations to learn a relation-oriented representation. Using the learned representation, we cluster unlabeled data and generate pseudo labels with higher accuracy. Based on the given labels and the generated pseudo labels, we optimize the model by minimizing a joint objective on both labeled and unlabeled data. Our experimental results on two real-world datasets show that our method outperforms current state-of-the-art methods for OpenRE.
The clustering-based unsupervised relation discovery method has gradually become one of the important methods of open relation extraction (OpenRE). 
However, high-dimensional vectors can encode complex linguistic information which leads to the problem that the derived clusters cannot explicitly align with the relational semantic classes.
In this work, we propose a relation-oriented clustering model and use it to identify the novel relations in the unlabeled data. Specifically, to enable the model to learn to cluster relational data, our method leverages the readily available labeled data of pre-defined relations to learn a relation-oriented representation. 
We minimize distance between the instance with same relation by gathering the instances towards their corresponding relation centroids to form a cluster structure, so that the learned representation is cluster-friendly.
To reduce the clustering bias on predefined classes, we optimize the model by minimizing a joint objective on both labeled and unlabeled data. Experimental results show that our method reduces the error rate by 29.2\% and 15.7\%, on two datasets respectively, compared with current SOTA methods.
\end{abstract}

\section{Introduction}
Relation extraction (RE), a crucial basic task in the field of information extraction, %automatically identifies the relations of a pair of entities in a sentence to produce a relational triplet. RE 
is of the utmost practical interest to various fields including web search \citep{10.1145/3038912.3052558}, knowledge base completion \citep{10.5555/2999792.2999923}, and question answering \citep{yu-etal-2017-improved}. However, conventional RE paradigms such as supervision and distant supervision are generally designed for pre-defined relations, which cannot deal with new emerging relations in the real world.
    \begin{figure}[t]
        \includegraphics[width=\columnwidth]{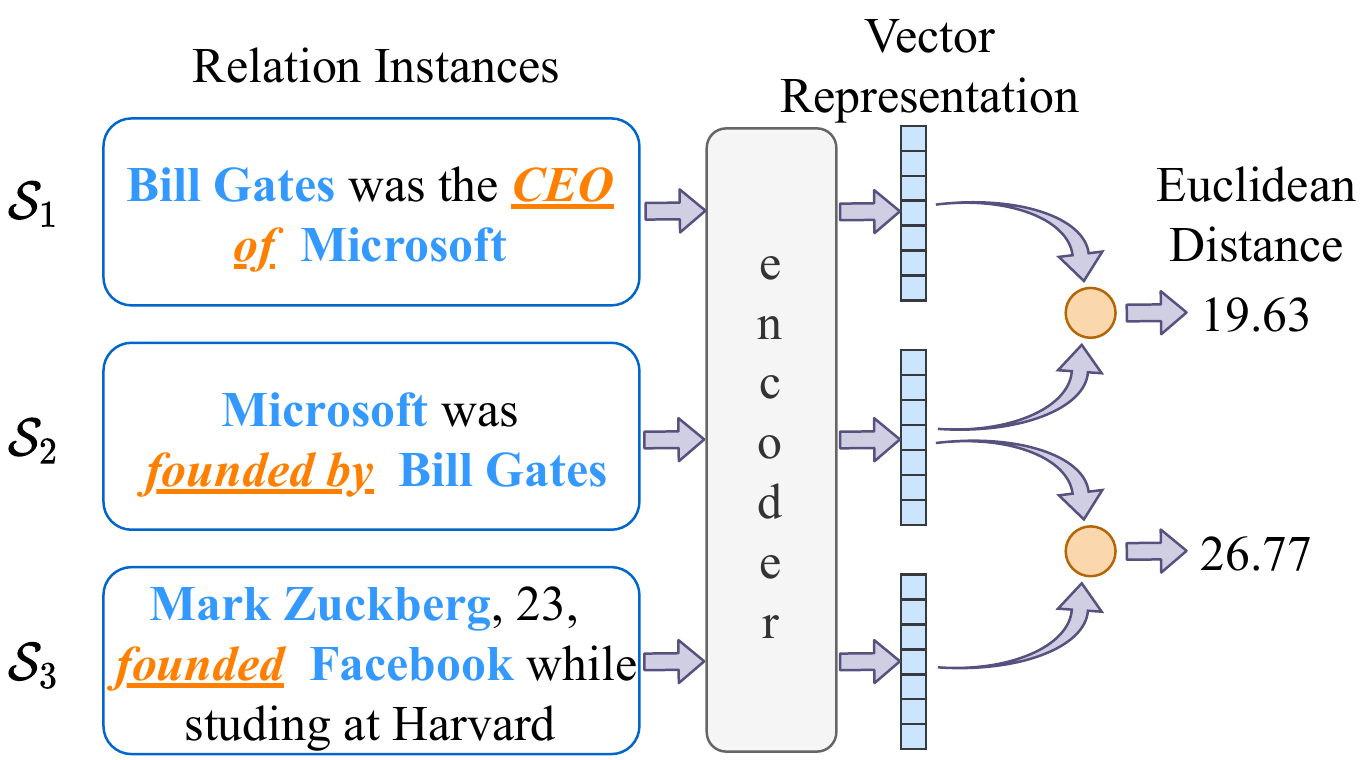}
        \caption{Although both instances $\mathcal{S}_2$ and $\mathcal{S}_3$ express \textit{founded} relation while $\mathcal{S}_1$ expresses \textit{CEO} relation, the distance between $\mathcal{S}_1$ and $\mathcal{S}_2$ is still smaller than that between $\mathcal{S}_2$ and $\mathcal{S}_3$. This is because there may be more similar surface information (e.g. word overlapping) or syntactic structure between $\mathcal{S}_1$ and $\mathcal{S}_2$, thus the derived clusters cannot explicitly align with relations.}
        \label{fig:intro}
    \end{figure}
    
    Under this background, open relation extraction (OpenRE) has been widely studied for its use in extracting new emerging relational types from open-domain corpora. The approaches used to handle open relations roughly fall into one of two groups. The first group is open information extraction (OpenIE) \citep{10.1145/1409360.1409378,yates-etal-2007-textrunner,fader-etal-2011-identifying}, which directly extracts related phrases as representations of different relational types. However, if not properly canonicalized, the extracted relational facts can be redundant and ambiguous. 
    %The second group,  unsupervised relation discovery \citep{yao-etal-2011-structured,shinyama-sekine-2006-preemptive,simon-etal-2019-unsupervised}, discovers unseen relation types from unsupervised data.
    The second group is unsupervised relation discovery \citep{yao-etal-2011-structured,shinyama-sekine-2006-preemptive,simon-etal-2019-unsupervised}.
    %In this type of research, much attention has been focused on unsupervised clustering-based RE methods, which cluster and recognize relation from high-dimensional representations. \citet{10.1007/978-3-319-70407-4_3} construct instance representation by re-weighting word embeddings and using NER tags as well as KB types as additional features. \citet{hu2020selfore} get the representation by concatenating the contextualized entity pair representation and self-supervised signals in pretrained language model are exploited for clustering optimization.
    In this type of research, much attention has been focused on unsupervised clustering-based RE methods, which cluster and recognize relations from high-dimensional representations \citep{10.1007/978-3-319-70407-4_3}. Recently, the self-supervised signals in pretrained language model are further exploited for clustering optimization \citep{hu2020selfore}.

    However, many studies show that high-dimensional embeddings can encode complex linguistic information such as morphological \citep{peters-etal-2018-dissecting}, local syntactic \citep{hewitt-manning-2019-structural}, and longer range semantic information \citep{jawahar-etal-2019-bert}. Consequently, the distance of representation is not completely consistent with relational semantic similarity. Although \citet{hu2020selfore} use self-supervised signals to optimize clustering, there is still no guarantee that the learned clusters will explicitly align with the desired relational semantic classes \citep{10.5555/2968618.2968683}. As shown in Figure \ref{fig:intro}, we use the method proposed by \citet{hu2020selfore} to get the instance representations. Although both instances $\mathcal{S}_2$ and $\mathcal{S}_3$ express the \textit{founded} relation, the euclidean distance between them is larger than that between $\mathcal{S}_1$ and $\mathcal{S}_2$, which express different relation. Obviously, the clustering algorithm tends to group instances $\mathcal{S}_1$ and $\mathcal{S}_2$ together, rather than $\mathcal{S}_2$ and $\mathcal{S}_3$ which express the same relation. %This is not desirable in our task.
    
    %a model can cluster relational instances according to surface information such as sentence length or local syntactic information such as type of entity pair. A systematic mechanism is needed to guide the model to cluster data in the desired way (e. g., cluster by relation).
    
    %In order to make the similarity of representation more accurately reflect the semantic similarity of the relations expressed in the instances, we propose a relation-oriented clustering method. In our proposed method, pre-defined relations and their existing labeled instances are leveraged to optimize a non-linear mapping, which transforms high-dimensional entity pair representations into relation-oriented representations. The transformed representations are scattered around their relational centroids, thereby the formation of clusters is based on relational information. In order to reduce the clustering bias on the predefined classes, we iteratively train the entity pair representations by optimizing a joint objective function on the labeled and unlabeled subsets of the data, improving both the supervised classification of the labeled data, and the clustering of the unlabeled data. In addition, the proposed method can be easily extended to incremental learning by classifying the pre-defined and novel relations with a unified classifier, which is often desirable in real-world applications. Our experimental results show that our proposed method outperforms current state-of-the-art methods for OpenRE.
    
    In this work, we propose a relation-oriented clustering method. To enable the model to learn to cluster relational data, pre-defined relations and their existing labeled instances are leveraged to optimize a non-linear mapping, which transforms high-dimensional entity pair representations into relation-oriented representations. Specifically, we minimize distance between the instances with same relation by gathering the instances representation towards their corresponding relation centroids to form the cluster structure, so that the learned representation is cluster-friendly. In order to reduce the clustering bias on the predefined classes, we iteratively train the entity pair representations by optimizing a joint objective function on the labeled and unlabeled subsets of the data, improving both the supervised classification of the labeled data, and the clustering of the unlabeled data. In addition, the proposed method can be easily extended to incremental learning by classifying the pre-defined and novel relations with a unified classifier, which is often desirable in real-world applications. Our experimental results show that our method outperforms current state-of-the-art methods for OpenRE. Our codes are publicly available at Github\footnote{https://github.com/Ac-Zyx/RoCORE.}.

    %Besides, we optimise the model by minimizing a joint objective function, containing terms for both the pre-defined and novel relations, using respectively the given labels and the generated constantly refined pseudo-labels, thus making the learned features are suitable for both the classification of pre-defined and novel relations. A further boost is obtained by incorporating incremental learning of the discovered relation types, which often desirable in real world applications.
    
    To summarize, the main contributions of our work are as follows: (1) we propose a novel relation-oriented clustering method RoCORE to enable model to learn to cluster relational data; %we propose a novel relation-oriented clustering method RoCORE for effectively guiding model clustering according to relation-related information; 
    (2) the proposed method achieves the incremental learning of unlabeled novel relations, which is often desirable in real-world applications; (3) experimental results show that our method reduces the error rate by 29.2\% and 15.7\%, on two real-world datasets respectively, compared with current state-of-the-art OpenRE methods. %our experimental results on two real-world dataset show that our method outperforms current state-of-the-art OpenRE methods and its performance is not sensitive to the number of types and samples of the pre-defined relations.
    \begin{figure*}[t]
    \centering
        \includegraphics[width=6.3in]{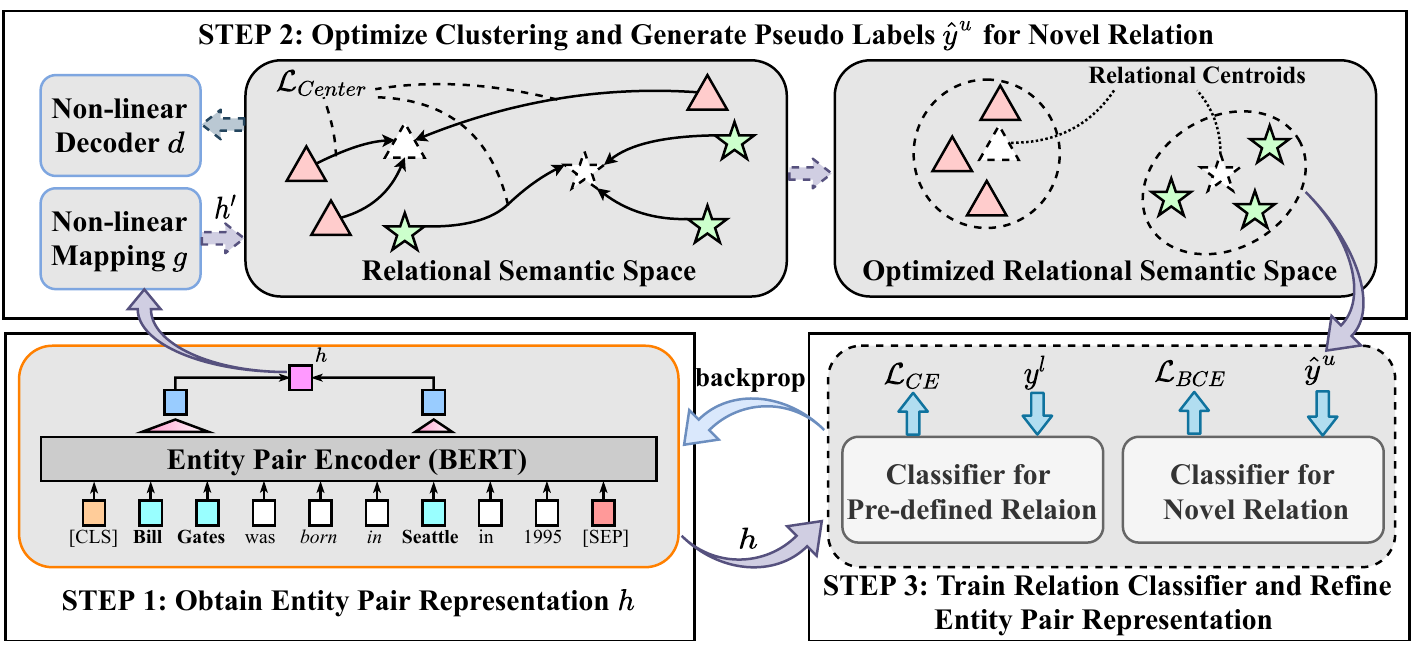}
        \caption{Overview of our RoCORE method. At the first step, we encode both the labeled and unlabeled instances in to entity pair representations. Then the entity pair representations are transformed to relation-oriented representations by gathering towards their relational centroids in the second step. Finally, based on the pseudo labels generated by clustering on unlabeled data, we optimize the entity pair representations and classifier by minimizing a joint objective function to reduce the clustering bias on predefined classes. The above three steps are performed iteratively to gradually improve model performance on novel relations.}
        \label{fig:model}
    \end{figure*}  
    
\section{Related Work}
    \textbf{Open Relation Extraction.} 
        To meet the needs of extracting new emerging relation types, many efforts have been undertaken to exploring methods for open relation extraction (OpenRE). The first line of research is Open Information Extraction \citep{10.1145/1409360.1409378,yates-etal-2007-textrunner,fader-etal-2011-identifying}, in which relation phrases are extracted directly to represent different relation types.
        However, using surface forms to represent relations results in an associated lack of generality since many surface forms can express the same relation. 
        Recently, unsupervised clustering-based RE methods is attracting lots of attentions. \citet{10.1007/978-3-319-70407-4_3} proposed to extract and cluster open relations by re-weighting word embeddings and using the types of named entities as additional features.  \citet{hu2020selfore} proposed to exploit weak, self-supervised signals in pretrained language model for adaptive clustering on contextualized relational features. However, the self-supervised signals are sensitive to the initial representation \citep{vangansbeke2020scan} and there is still no guarantee that the learned clusters will align with the relational semantic classes \citep{10.5555/2968618.2968683}. %To the best of our knowledge, the work most related to ours is 
        \citet{wu-etal-2019-open} proposed the relation similarity metrics from labeled data, and then transfers the relational knowledge to identify novel relations in unlabeled data. Different from them, we propose a relation-oriented method explicitly clustering data based on relational information.
        
    \noindent\textbf{Knowledge in High-Dimensional Vector.}
        Pre-trained static and contextual word representations can provide valuable prior knowledge for constructing relational representations \citep{DBLP:journals/corr/abs-1906-03158,10.1007/978-3-319-70407-4_3}. % and significant progress has been made in relation extraction tasks  However, high-dimensional vectors can encode complex linguistic information ranging from syntax to semantics.
        \citet{peters-etal-2018-dissecting} showed that different neural architectures (e.g., LSTM, CNN, and Transformers) can hierarchically structure linguistic information %, such that morphological, local syntactic, and longer range semantic information tend to be encoded into the high-dimensional embedding 
        that varies with network depth. 
        Recently, many studies \citep{jawahar-etal-2019-bert,DBLP:journals/corr/abs-1906-04341,DBLP:journals/corr/abs-1901-05287} have shown that such hierarchy also exists in pretraining models like BERT. %\citet{hewitt-manning-2019-structural} showed that we can recover parse trees from the linear transformation of contextual word representation consistently, better than with non-contextual baselines. These results suggest that high-dimensional embeddings, independent of model architecture, are learning much more about the structure of language than previously appreciated.
        These results suggest that  high-dimensional embeddings, independent of model architecture, learn much about the structure of language. Directly clustering on these high-dimensional embeddings should hardly produce ideal clusters in our desired way, which motivates us to extend current unsupervised clustering-based RE methods to learn the representations tailored for clustering relational data.
        
\section{Approach}
    In this work, we propose a relation-oriented clustering method, which takes advantage of the relational information in the existing labeled data to enable model to learn to cluster relational data. In order to reduce the clustering bias on the predefined classes, we iteratively train the entity pair representations by optimizing a joint objective function on the labeled and unlabeled subsets of the data, improving both the supervised classification of the labeled data, and the clustering of the unlabeled data. The proposed method is shown in Figure \ref{fig:model}.
    
    Specifically, %we consider the problem of relation discovery from unsupervised data, taking advantage of readily available labeled data of pre-defined relation. Formally, 
    given an unlabeled dataset $\mathcal{D}^u=\{s_i^u\}_{i=1,...,M}$ of relational instances $s_i^u$, our goal is to automatically cluster the relational instances into a number of classes $C^u$, which we assume to be known a priori. To enable the model to learn to cluster data, we incorporate a second labeled dataset of pre-defined relations $\mathcal{D}^l=\{(s_i^\ell,y_i^\ell)\}_{i=1,...,N}$ where $y_i^\ell\in \{1,...,C^\ell\}$ is the relational label for instance $s_i^\ell$.
    \subsection{Method Overview}
        We approach the problem by learning a relation-oriented representation, from which the derived clusters can be explicitly aligned with the desired relational semantic classes.
        %which is suitable for clustering relation-related data. 
        As illustrated in Figure \ref{fig:model}, we learn the representation and optimize the model by performing three iterative steps:

        \noindent(1) First, we encode relation instances in $\mathcal{D}^\ell$ and $\mathcal{D}^u$ using the entity pair encoder implemented as the pretrained BERT \citep{DBLP:journals/corr/abs-1810-04805}, which takes relation instances $\{s_i^\ell\}_{i=1,...,N}$, and $\{s_j^u\}_{j=1,...,M}$, as input, and output relation representation $\bm{h_i^\ell}$, $\bm{h_j^u}$. However, high-dimensional $\bm{h}$ can encode a mixture of various aspects of linguistic features and the derived clusters from $\bm{h}$ cannot explicit align with desired relational classes. %distance between $\bm{h}$ is not completely consistent with relational semantic similarity.

        \noindent(2) To make the distance between the representations accurately reflect the relational semantic similarity, the obtained $\bm{h}_i^\ell$ are transformed to low-dimensional relation-oriented representations $\bm{h}_i^{\ell\prime}$ by a non-linear mapping $\bm{g}$. Under the supervision of labels $y_i^\ell$ in $\mathcal{D}^\ell$, $\bm{g}$ is optimized by the gathering of $\bm{h}_i^{\ell\prime}$ towards their relational centroids to form a cluster structure, thereby we obtain $\bm{h}_j^{u\prime}$ from unlabeled data using the optimized $\bm{g}$ and generate the pseudo labels $\hat{y}^u$ according to clustering on $\bm{h}_i^{u\prime}$. 

        \noindent(3) Because using labeled data to guide the $\bm{h^\prime}$ towards their relational centroids will produce clustering bias on pre-defined relations, it is difficult to directly generate high-quality pseudo labels. To reduce the negative effect of errors in pseudo labels, we optimize classifier and entity pair representations by minimizing a joint objective function, containing terms for both pre-defined and novel relations, using respectively the given labels $y^\ell$ and generated pseudo label $\hat{y}^u$. Based on the refined entity pair representation $\bm{h}$ which encode more contextual relational information, the above three steps are performed iteratively to gradually improve the quality of pseudo labels $\hat{y}^u$ and model performance.
        
        \subsection{Entity Pair Encoder}
        Given a relation instance $s_i=(\bm{x}_i,h_i, t_i)$, which consists of a sentence $\bm{x}_i=\{x_1,x_2,...,x_n\}$ and two entity spans $h_i=(s_h,e_h)$, $t_i=(s_t,e_t)$ marking the position of the entity pair, the entity pair encoder $\bm{f}$ aims to map relation instance $s_i$ to a fixed-length embedding $\bm{h}_i=\bm{f}(s_i) \in \mathbb{R}^d$ that encode contextual information in $s_i$. We adopt BERT \citep{DBLP:journals/corr/abs-1810-04805} as the implemention of our encoder $\bm{f}$ due to its strong performance on extracting contextual information. Formally:
        \begin{gather}
            \bm{h}_1^r,...,\bm{h}_n^r={\rm BERT}^r(x_1,...,x_n)\\
            \bm{h}_{ent}={\rm MAXPOOL}(\bm{h}_s^r,...,\bm{h}_e^r)\\     
            \bm{h}_i=\bm{h}_{head}\oplus\bm{h}_{tail},
        \end{gather}
        where $r$ is a hyperparameter that denotes the output layer of BERT. $s$ and $e$ represent start and end position of the corresponding entity respectively. $\oplus$ denotes the concatenation operator. This structure of entity pair representation encoder has been widely used in previous RE methods \citep{wang-etal-2021-enpar,hu2020selfore}.

        \subsection{Relation-Oriented Clustering Module}
            %In order to cluster data according to relational features, 
            In order to make the distance between representation accurately reflect the relational semantic similarity, the obtained $\{\bm{h}^\ell_i\}_{i=1,...,N}$ are transformed to low-dimensional relation-oriented representations $\bm{h}_i^{\ell\prime}$ by a non-linear mapping $\bm{g}(\cdot):\mathbb{R}^d\rightarrow\mathbb{R}^m$. Under the supervision of labels $y_i^\ell$ in $\mathcal{D}^\ell$, $\bm{g}$ is optimized by the gathering of $\bm{h}_i^{\ell\prime}$ towards their relational centroids as follows:
            \begin{gather}
                \mathcal{L}_{center}=\frac{1}{2N}\sum_{i=1}^{N}\left\|\bm{h^{\ell\prime}}_i-\bm{c}_{y_i}\right\|_2^2\\
                \bm{c}_{r}=\frac{1}{|\mathcal{D}_r|}\sum_{i\in \mathcal{D}_r}\bm{h}_i^{\ell\prime},
            \end{gather}
            where $\bm{c}_r$ denotes the centroids of relation $r$. The center loss $\mathcal{L}_{center}$ seems reasonable, but problematic. A global optimal solution to minimize $\mathcal{L}_{center}$ is $g(\bm{h}_i)=\bm{0}$, which is far from being desired. This motivates us to incorporate a reconstruction term to prevent the semantic space from collapsing. Specifically,, a decoding network $\bm{d}(\cdot)$ is used to map the representation $\bm{h}_i^{\prime}$ back to the original representation $\bm{h}_i$.Thus, we can derive the following loss function:
            \begin{equation}
                \mathcal{L}_{C}=\frac{1}{2N}\sum_{i=1}^{N}\ell(\bm{d}(\bm{h}_i^\prime),\bm{h}_i)+\lambda\mathcal{L}_{center},
            \end{equation}
            where both the encoder $\bm{g}(\bm{h}_i)$ and decoder $\bm{d}(\bm{h}_i^\prime)$ are implemented as DNN. The function $\ell(\cdot,\cdot):\mathcal{R}^d\rightarrow \mathcal{R}$ is the least-squares loss $\ell(\bm{x},\bm{y})=\left\|\bm{x}-\bm{y}\right\|_2^2$ that measures the reconstruction error and other choices such as $\ell_1$-norm also can be considered. $\lambda$ is a hyper-parameter that balances the reconstruction error versus center loss.

            Finally, we obtain $\{\bm{h}_j^{u\prime}\}_{j=1,...,M}$ using the optimized $\bm{g}$ and generate pseudo labels $\hat{y}^u$ using k-means algorithm as follows:
            \begin{equation}
                \hat{y}^u=\text{ k-means}(\bm{h}^{u\prime}), \label{equ:plabel}
            \end{equation}
            
        \subsection{Relation Classification Module}
        Based on the pseudo labels $\hat{y}^u$ generated by clustering, we can train the classifier and refine entity pair representation $\bm{h}$ to encode more contextual relation information. Since it’s difficult to keep the order of clusters consistent in multiple clustering, instead of using standard cross entropy loss, we propose to use the pairwise similarities for novel relation learning. 
            \begin{equation}
                q_{ij}=\mathbbm{1}\{\hat{y}_i^u=\hat{y}_j^u\},
            \end{equation}
        where the symbol $q_{ij}$ denotes whether $s_i^u$ and $s_j^u$ belong to the same cluster. If a pair is from the same cluster, the classifier $\bm{\eta}^u:\mathcal{R}^d\rightarrow\mathcal{R}^{C^u}$ outputs similar distributions, and vice-versa. Specifically, we use the pair-wise KL-divergence to evaluate the distance of two relation instances. Given a pair of instance $s_i^u$, $s_j^u$, their corresponding output distributions are defined as $\mathcal{P}=\bm{\eta}^u(\bm{f}(s_i^u))$ and $\mathcal{Q}=\bm{\eta}^u(\bm{f}(s_j^u))$. For the pair from the same cluster, the cost is described as:
            \begin{gather}
                \mathcal{L}^+(s_i^u,s_j^u)=\mathcal{D}_{KL}(\mathcal{P^*}||\mathcal{Q})+\mathcal{D}_{KL}(\mathcal{Q^*}||\mathcal{P})\\
                \mathcal{D}_{KL}(\mathcal{P^*}||\mathcal{Q})=\sum_{c=1}^{C^u}p_clog\frac{p_c}{q_c},
            \end{gather}
        where $\mathcal{P}^*$ denotes that $\mathcal{P}$ is assumed to be a constant and each KL-divergence factor $\mathcal{D}_{KL}(\mathcal{P}||\mathcal{Q})$ is a unary function whose gradient is simply $\partial\mathcal{D}_{KL}(\mathcal{P}^*||\mathcal{Q})/\partial\mathcal{Q}$. 
        
        If $s_i^u$, $s_j^u$ comes from different clusters, their output distributions are expected to be different, which can be defined as a hinge-loss function:
            \begin{gather}
                \begin{split} 
                    \mathcal{L}^-(s_i^u,s_j^u)=L_h(\mathcal{D}_{KL}(\mathcal{P^*}||\mathcal{Q}),\sigma)+\\
                    L_h(\mathcal{D}_{KL}(\mathcal{Q^*}||\mathcal{P}),\sigma)
                \end{split}
               \\
                L_h(e,\sigma)=\max (0,\sigma-e),
            \end{gather}            
            and the total loss can be defined as a contrastive loss:
            \begin{equation}
                \begin{split}
                    \mathcal{L}_{BCE}(s_i^u,s_j^u)=q_{i,j}\mathcal{L}^+(s_i^u,s_j^u)+
                    \\(1-q_{ij})\mathcal{L}^-(s_i^u,s_j^u).
                \end{split}
            \end{equation}
        Note that $\mathcal{L}_{BCE}$ is a symmetric loss w.r.t. $s_i^u$,$s_j^u$ since $\mathcal{P}$ and $\mathcal{Q}$ are alternatively assumed to be constant in $\mathcal{L}^+$ and $\mathcal{L}^-$. %Each KL-divergence factor $\mathcal{D}_{KL}(\mathcal{P}||\mathcal{Q})$ becomes a unary function whose gradient is simply $\partial\mathcal{D}_{KL}(\mathcal{P}^*||\mathcal{Q})/\partial\mathcal{Q}$.
        %Since we set the output dimension of $\bm{\eta}^u$ to be equal to the number of new classes $C^u$, we can use the index of the maximum element of the output distribution as the prediction $\hat{y}_i^u$ for the relation instance $s_i^u$:
        Finally, we get the prediction for a relation instance $s_i^u$ as follows:
        \begin{equation}
            \hat{y}_i^u=\arg \max_y[\bm{\eta}^u(\bm{f}(s_i^u))]_y    \label{equ:yu}
        \end{equation}
    \subsection{Training Methods}  
        \subsubsection{Iterative Joint Training}
        Because using labeled data to guide the $\bm{h}$ towards their relational centroids will produce clustering bias on pre-defined relations, it is difficult to directly generate high-quality pseudo labels $\hat{y}^u$ for novel relations. To reduce the negative effect of errors in pseudo labels, we incorporate a classifier $\bm{\eta}^\ell:\mathcal{R}^d\rightarrow\mathcal{R}^{C^l}$ for pre-defined relations and refine $\bm{h}$ by minimizing a joint objective function, containing terms for both pre-defined and novel relations, using respectively the given labels $y^\ell$ and generated pseudo label $\hat{y}^u$ as follows:
        \begin{gather}
          \mathcal{L}_{CE}=-\frac{1}{N}\sum_{i=1}^{N}log\bm{\eta}^\ell_{y_i}(\bm{h_i}) \label{equ:CE}\\
          \mathcal{L}_{CLS}=\mathcal{L}_{CE}+\mathcal{L}_{BCE}.
        \end{gather}

        % \noindent\textbf{Iterative Optimization for Boosting Performance}.
        The refined entity pair representation $\bm{h}$ encode more contextual relation information, which in turn promote clustering optimization and generate pseudo labels $\hat{y}^u$ with higher accuracy.
        We refine representation $\bm{h}$ and optimize clustering in a iterative manner to gradually improve the quality of the pseudo labels and model performance. This iterative procedure is detailed in Algorithm \ref{alg:1}.

            \begin{algorithm}
                \caption{The RoCORE Method}
                \label{alg:1}
                \KwIn{
                    novel relation dataset $\mathcal{D}^u=\{s_j^u\}$,\\ predefined relation dataset $\mathcal{D}^\ell=\{(s_i^\ell,y_i^\ell)\}$, \\model parameters $\Theta$, $\Phi$, $\Psi$ for Entity pair encoder, Relation-oriented clustering module, Relation classifiers, respectively,\\
                    and learning rate $\eta$.
                }
                \For{$epoch\leftarrow 1$ \KwTo $L$}{
                    Pre-train clustering network by minimize reconstruction loss $\Phi=\Phi - \eta\nabla_{\Phi} \ell(\bm{g}(\bm{h}_i^\prime),\bm{h}_i)$\;
                }
                \Repeat{\text{convergence}}{
                    generate pseudo labels $\hat{y}$ by equation \ref{equ:plabel}\;
                    refine entity pair representation\\
                    $\Theta=\Theta - \eta\nabla_{\Theta} \mathcal{L}_{CLS}$\;
                    $\Psi=\Psi - \eta\nabla_{\Psi} \mathcal{L}_{CLS}$\;
                    optimize clustering
                    $\Phi=\Phi - \eta\nabla_{\Phi}\mathcal{L}_{C}$\;
                }
            \end{algorithm}

    \subsubsection{Incremental Learning Scheme} In real-world settings, when facing a new sentence, we often don't know whether it belongs to pre-defined relations or novel relations. In this work, we explore the incremental learning of novel relations to enable $\bm{\eta}^l$ to discriminate both pre-defined and novel relations. Under incremental learning settings, we extend the classifier $\bm{\eta}^l$ to $C^u$ novel relation types, so that $\bm{\eta}^l:\mathcal{R}^d\rightarrow\mathcal{R}^{C^l+C^u}$. Then, the model is trained using cross-entropy loss instead of equation \ref{equ:CE} as follows:
            \begin{equation}
               \begin{aligned} \mathcal{L}_{CE}=&-\frac{1}{N}\sum_{i=1}^{N}log\bm{\eta}^\ell_{y_i}(\bm{h}_i)\\&-\frac{\mu(t)}{M}\sum_{j=1}^{M}log\bm{\eta}^\ell_{\hat{y}_j}(\bm{h}_j),
               \end{aligned}
            \end{equation}
    where we obtain $\hat{y}_j$ using equation \ref{equ:yu} and the coefficient $\mu(t)$ balances the cross entropy loss of pre-defined and novel relations. We implemented it as a ramp-up function $\mu(t)=\mu_0e^{-5(1-\frac{t}{T})^2}$ where $t$ is current epoch and $T$ is the ramp-up length and coefficient $\mu_0\in \mathbb{R}^+$.

\section{Experimental Setup}
    \label{sec:exp_setup}
    In this section, we describe the datasets for training and evaluating the proposed method. We also detail the baseline models
    %and evaluation metrics 
    for comparison.  Finally, we clarify the implementation details and hyperparameter configuration of our method.
    \subsection{Datasets}
        We conduct experiments on two relation extraction datasets.
        
        \noindent\textbf{FewRel}. Few-Shot Relation Classification Dataset \citep{han-etal-2018-fewrel}. FewRel is a human-annotated dataset containing 80 types of relations, each with 700 instances. We follow the setting in \citep{wu-etal-2019-open} to use the original train set of FewRel, which contains 64 relations, as labeled set with predefined relations, and the original validation set of FewRel, which contains 16 new relations, as the unlabeled set with novel relations to extract. 1,600 instances were randomly selected from the unlabeled set as the test set. The rest of labeled and unlabeled instances are considered as the train set.
        
        %\noindent\textbf{FewRel-distant}. The second dataset we use is FewRel-distant, which contains the distantly-supervised data obtained by the authors of FewRel before human annotation.
        
        \noindent\textbf{TACRED}. The TAC Relation Extraction Dataset \citep{zhang-etal-2017-position}. TACRED is a human-annotated large-scale relation extraction dataset that covers 41 relation types. We remove the instances labeled as \texttt{no\_relation} and use the remaining 21,773 instances for training and evaluation. Similar to the setting of FewRel, we select the 0-30 relation types as labeled set with pre-defined relations and the 31-40 relation types as unlabeled set with novel relations. We randomly selected 15\% of the instances from the unlabeled set as the test set. The rest of the labeled and unlabeled instances are considered as the train set.
        
    \subsection{Compared Methods}% and Evaluation Metrics}
        To evaluate the effectiveness of our method, we select the following SOTA OpenRE models for comparison. Note that the first four methods are unsupervised and RSN as well as RSN-BERT leverages labeled data of predefined relations.
        
        \noindent\textbf{HAC with Re-weighted Word Embeddings (RW-HAC)} \citep{10.1007/978-3-319-70407-4_3}. RW-HAC is a feature clustering method for OpenRE. The model contructs relational feature based on the weighted word embeddings as well as entity types.
        %types of the involved named entities and the weighted sum of pretrained word embeddings according to whether they are terms forming the relations. %After dimension reduction by principal component analysis (PCA), the concatenation of reduced feature will be clustered by Hierarchical Agglomerative Clustering (HAC).
        
        \noindent\textbf{Discrete-state Variational Autoencoder (VAE)} \citep{marcheggiani-titov-2016-discrete}. VAE is a reconstruction-based method for OpenRE. The model is optimized by reconstructing entities from pairing entities and predicted relations. % types. Features such as pos, trigger words, dependency path are used to predict relations.%Rich features including entity and trigger words, dependency paths, and POS tags are used to predict the relation type.
        
        \noindent\textbf{Entity Based URE (Etype+)} \citep{tran-etal-2020-revisiting}. Etype+ is a simple and effective method relying only on entity types. The same link predictor as in \citep{marcheggiani-titov-2016-discrete} is employed and two additional regularisers are used.
        
        \noindent\textbf{Self-supervised Feature Learning for OpenRE (SelfORE)} \citep{hu2020selfore}. SelfORE exploits weak, self-supervised signals by leveraging large pretrained language model for adaptive clustering on contextualized relational features.
        
        \noindent\textbf{Relational Siamese Network (RSN)} \citep{wu-etal-2019-open}. This method learns similarity metrics of relations from labeled data of pre-defined relations, and then transfer the relational knowledge to identify novel relations in unlabeled data. 
        
        \noindent\textbf{RSN with BERT Embedding (RSN-BERT)}. A variant of RSN, the static word vector is replaced by the BERT embedding for fair comparison.
        
        %In evaluation, we use B3 metric \citep{Bagga1998AlgorithmsFS} as the scoring function. B3 metric is a standard measure to balance the precision and recall of clustering tasks, and is commonly used in previous OpenRE works \citep{marcheggiani-titov-2016-discrete,10.1007/978-3-319-70407-4_3,wu-etal-2019-open} To be specific, we use F1 measure, the harmonic mean of precision and recall.
        
    \subsection{Implementation Details}
        Our entity pair encoder is implemented as the \texttt{ bert-base-uncased} which consists of 12 layers and we use layer 8 as the output layer for best performance. Note that we only fine-tune the parameters of the output layer in the iterative training process to avoid overfitting. Non-linear mapping $\bm{g}(\cdot)$ and $\bm{d}(\cdot)$ are both implemented as a DNN with \texttt{relu} activation, specifically $\mathbb{R}^d$-512-512-256 for $\bm{g}(\cdot)$ and 256-512-512-$\mathbb{R}^d$ for $\bm{d}(\cdot)$. All experiments are conducted using a GeForce GTX 1080Ti with 11GB memory and table \ref{tab:hyper} shows our best hyper-parameter settings.
        \begin{table}
            \centering
            \begin{tabular}{lr}
            \toprule
            Hyper-parameters & value\\
            \midrule
            optimizer & \textit{Adam}\\
            learning rate & 1e-4\\
            batch size & 100\\
            \midrule
            pre-training epochs $L$ & 10\\
            BCE loss coefficient $\sigma$ & 2\\
            center loss coefficient $\lambda$ for FewRel & 0.005\\
            center loss coefficient $\lambda$ for TACRED & 0.001\\
            ramp-up coefficient $\mu_0$ & 1.0\\
            ramp-up length $T$ & 10\\
            \bottomrule
            \end{tabular}
            \caption{Hyper-parameter settings.}
            \label{tab:hyper}
        \end{table}
        \begin{table*}
            \centering
            \resizebox{\linewidth}{!}{
            \begin{tabular}{ll lll lll l}
            \toprule
            \multirow{2}{*}{\textbf{Dataset}} & \multirow{2}{*}{\textbf{Method}} & \multicolumn{3}{c}{$B^3$} & \multicolumn{3}{c}{V-measure} & \multirow{2}{*}{ARI}\\
            \cline{3-8}
            & & Prec. & Rec. & $F_1$ & Hom. & Comp. & $F_1$\\
            \midrule
            \multirow{7}{*}{\textbf{FewRel}} %& VAE & 0.348 & 0.498 & 0.410 & 0.484 & 0.535 & 0.508 & 0.341\\
            & VAE\citep{marcheggiani-titov-2016-discrete} & 0.309 & 0.446 & 0.365 & 0.448 & 0.500 & 0.473 & 0.291\\
            &RW-HAC\citep{10.1007/978-3-319-70407-4_3} & 0.256 & 0.492 & 0.337 & 0.391 & 0.485 & 0.433 & 0.250 \\
            %&UIE \citep{simon-etal-2019-unsupervised}& 0.191 & 0.258 & 0.220 & 0.303 & 0.314 & 0.308 & 0.172\\
            &EType+\citep{tran-etal-2020-revisiting} & 0.238 & 0.485 & 0.319 & 0.364 & 0.463 & 0.408 & 0.249 \\
            &SelfORE\citep{hu2020selfore} & 0.672 & 0.685 & 0.678 & 0.779 & 0.788 & 0.783 & 0.647\\
            \cline{2-9}
            &RSN\citep{wu-etal-2019-open}& 0.486 & 0.742 & 0.589 & 0.644 & 0.787 & 0.708 & 0.453\\
            &RSN-BERT& 0.585 & 0.899 & 0.709 & 0.696 & 0.889 & 0.781 & 0.532\\
            &\textbf{RoCORE}& $0.752_{17}$ & $0.846_{09}$ & $\textbf{0.796}_{11}$ & $0.838_{10}$ & $0.883_{06}$ & $\textbf{0.860}_{07}$ & $\textbf{0.709}_{23}$ \\
            \hline \hline
            \multirow{7}{*}{\textbf{TACRED}} & VAE\citep{marcheggiani-titov-2016-discrete}& 0.247 & 0.564 & 0.343 & 0.208 & 0.362 & 0.264 & 0.159 \\
            %&RW-HAC & 0.362 & 0.888 & 0.514 & 0.366 & 0.772 & 0.497 & 0.224 \\
            %&RW-HAC\citep{10.1007/978-3-319-70407-4_3} & 0.684 & 0.754 & 0.718 & 0.727 & 0.762 & 0.744 & 0.665 \\
            &RW-HAC\citep{10.1007/978-3-319-70407-4_3} & 0.426 & 0.633 & 0.509 & 0.469 & 0.597 & 0.526 & 0.281 \\
            %&UIE \citep{simon-etal-2019-unsupervised}& 0.242 & 0.193 & 0.215 & 0.202 & 0.184 & 0.192 & 0.074\\
            &EType+\citep{tran-etal-2020-revisiting} & 0.302 & 0.803 & 0.439 & 0.260 & 0.607 & 0.364 & 0.143\\
            &SelfORE\citep{hu2020selfore} & 0.576 & 0.510 & 0.541 & 0.630 & 0.608 & 0.619 & 0.447\\
            \cline{2-9}
            &RSN\citep{wu-etal-2019-open}& 0.628 & 0.634 & 0.631 & 0.624 & 0.663 & 0.643 & 0.459 \\
            &RSN-BERT& 0.795 & 0.878 & 0.834 & 0.849 & 0.870 & 0.859 & 0.756\\
            &\textbf{RoCORE}& $0.871_{42}$ & $0.849_{37}$ & $\textbf{0.860}_{35}$ & $0.895_{32}$ & $0.881_{20}$ & $\textbf{0.888}_{24}$ & $\textbf{0.812}_{64}$\\
            \bottomrule
            \end{tabular}
            }
            \caption{Main results on two relation extraction datasets. The subscript represents the corresponding standard deviation (e.g., $0.796_{11}$ indicates $0.796\pm0.011$). Experimental results show that our method reduces the error rate by 29.2\%(0.709$\rightarrow$0.796) and 15.7\%(0.834$\rightarrow$0.860), on two datasets respectively.} %It is worth noting that when a model tends to cluster multiple relation into one, the produced results will have a high recall and a low precision value, which is undesired in many real-world applications. RSN-BERT in this table is an example.}
            \label{tab:main_res}
        \end{table*}           
\section{Results and Analysis}   
    In this section, we present the experimental results of our model on two real-world datasets % and comprehensive analyses 
    to demonstrate the effectiveness of our method. We also provide additional experimental results on hyper-parameter analysis and relation representation visualization in appendix \ref{app:Hyper} and \ref{app:visual}.%\textbf{A} and \textbf{B}.
    \subsection{Main Results}
        
        Table \ref{tab:main_res} reports model performances on FewRel, TACRED dataset, which shows that the proposed method achieves state-of-the-art results on OpenRE task. %Benefitting from the valuable information in the labeled instances of pre-defined relations, RoCORE effectively learns the relation-oriented representation which are suitable for clustering relational data, thereby outperforming previous unsupervised clustering-based baseline RW-HAC and SelfORE by a large margin. 
        Benefitting from the valuable information in the labeled instances of pre-defined relations, RoCORE effectively learns the relation-oriented representation from which the derived clusters explicitly align with relational semantic classes, thereby outperforming previous clustering-based baseline such as SelfORE by a large margin. 
        In addition, despite the fact that RSN and its variant RSN-BERT also leverage relational information in labeled data, 
        %pre-defined relational dataset, 
        the learning of similarity metrics and clustering are mutually independent. In our method, relation representation learning and cluster optimization are mutually dependent. Thus, the learned representations are tailored for clustering. As a result, our method outperforms RSN and RSN-BERT on the two datasets.

    \subsection{Ablation Study}
        \label{sec:ablation}
        \begin{table}
            \centering
            \resizebox{\columnwidth}{!}{
            \begin{tabular}{ll lll}
            \toprule
            \textbf{Dataset} & \textbf{Method} & Prec. & Rec. & $F_1$\\
            \midrule
            \multirow{4}{*}{\textbf{FewRel}}
            &w/o center loss & $0.726_{32}$ & $0.774_{31}$ & $0.749_{31}$\\
            &w/o reconstruction & $0.512_{38}$ & $0.573_{17}$ & $0.540_{25}$\\
            &w/o CE & $0.662_{67}$ & $0.787_{54}$ & $0.719_{47}$\\
            &\textbf{RoCORE}& $0.752_{17}$ & $0.846_{09}$ & $\textbf{0.796}_{11}$\\
            \hline
            \multirow{4}{*}{\textbf{TACRED}}
            &w/o center loss & $0.818_{57}$ & $0.842_{37}$ & $0.830_{41}$\\
            &w/o reconstruction & $0.549_{35}$ & $0.483_{30}$ & $0.514_{31}$\\
            &w/o CE & $0.706_{45}$ & $0.776_{51}$ & $0.739_{47}$\\
            &\textbf{RoCORE}& $0.871_{42}$ & $0.849_{37}$ & $\textbf{0.860}_{35}$\\
            \bottomrule
            \end{tabular}
            }
            \caption{Abalation study of our method. This table only lists the results of metric $B^3$. For results of other metrics, please refer to the Table \ref{tab:det_aba} in Appendix \ref{app:other}.}
            \label{tab:abalation}
        \end{table}              
        To study the contribution of each component in the proposed method, we conduct ablation experiments on the two datasets and display the results in Table \ref{tab:abalation}. The results show that the model performance is degraded if $\mathcal{L}_{center}$ is removed, indicating that the guidance of supervision signals from pre-defined relations provide valuable information for learning the relation-oriented representations. It is worth noting that the reconstruction term has an important role in the clustering module. Without the reconstruction term, the semantic space will collapse and the performance will be seriously hurt. In addition, joint optimizing on both the labeled and unlabeled data is also very important. The initial pseudo labels for novel relations are not accurate due to the unwanted clustering bias on pre-defined relations. Without $\mathcal{L}_{CE}$, the error in pseudo labels will lead the refinement of the entity pair representation to a wrong direction, which affects the model performance.

    \subsection{The Influence of Pre-defined Relation Number on Performance}
        \begin{figure}[t]
            \includegraphics[width=\columnwidth]{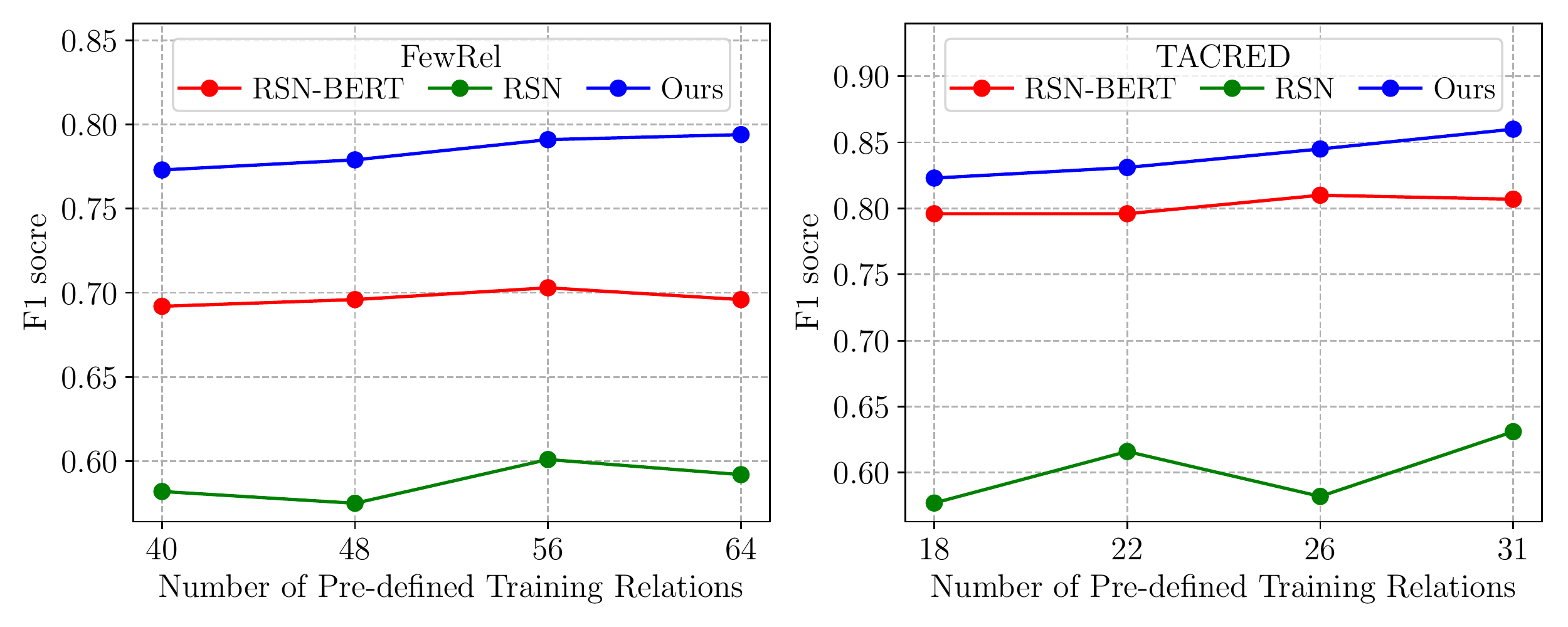}
            \caption{Clustering results with different numbers of pre-defined training relations.}
            \label{fig:num_class}
        \end{figure}
        In this subsection, we conduct experiments on two different datasets to explore the influence of pre-defined relation number on performance of our method. For FewRel dataset, following the setting in \citep{wu-etal-2019-open}, we change the number of pre-defined relations from 40 to 64 while fixing the total number of labeled instances to 25,000. Similarly, the settings for TACRED dataset is 18, 31 and 12, 000, respectively.
        %For TACRED dataset, the number of pre-defined relations are changed from 18 to 31 while fixing the total number of labeled instances to 12,000. 
        
        %The results on the two different datasets show similar patterns:
        From figure \ref{fig:num_class} we can see the following: (1) The increase of pre-defined relation number do improve the generalization of our method on novel relations. The models trained on 64/31 relations slightly perform better than the models trained on 40/18 relations on FewRel/TACRED dataset (2) Our method constantly performs better than RSN and RSN-BERT with the number of predefined relations vary. This indicates the effectiveness of our method.
        
    \subsection{Cross Domain Analysis}
        \begin{table}
            \centering
            \resizebox{\linewidth}{!}{
            \begin{tabular}{ll lll}
            \toprule
            \textbf{Task} & \textbf{Method} & Prec. & Rec. & $F_1$\\
            \midrule
            \multirow{3}{*}{$\textbf{F}\rightarrow \textbf{T}$}
            &RSN & 0.349 & 0.590 & 0.439\\
            &RSN-BERT & 0.337 & 0.866 & 0.486\\
            &\textbf{RoCORE} & $0.621_{28}$ & $0.602_{51}$ & $\textbf{0.611}_{34}$\\
            \hline
            \multirow{3}{*}{$\textbf{T}\rightarrow\textbf{F}$}
            &RSN & 0.225 & 0.529 & 0.316 \\
            &RSN-BERT & 0.261 & 0.861 & 0.400\\
            &\textbf{RoCORE}& $0.687_{36}$ & $0.766_{46}$ & $\textbf{0.724}_{26}$\\
            \bottomrule
            \end{tabular}
            }
            \caption{\label{tab:cross_res}
            Results on two constructed cross-domain tasks. F means FewRel, which is from encyclopedia domain. T means TACRED, which is from news and web domain. This table only lists the results of metric $B^3$. For results of other metrics, please refer to the Table \ref{tab:det_cro} in Appendix \ref{app:other}.%\textbf{C}.
            }
        \end{table}   
        
        In real-world settings, pre-defined relations and novel relations of interest usually come from different domains. To study the model performance in cross-domain settings, we conducted experiments on two cross-domain tasks, i.e,: FewRel to TACRED and TACRED to FewRel. Pre-defined relations and their labeled instances come from the source domain training dataset, and we evaluate performance on the target domain testing dataset. 
        
        Table \ref{tab:cross_res} shows the experimental results, from which we can observe that: %(1) unsupervised relation discovering methods are not affected because they do not use labeled data from the source domain;
        (1) the change of domain increases the semantic gap between the pre-defined and novel relations. As the result of that, the performance of the model using labeled data of predefined relations is degraded. (2) compared with RSN and RSN-BERT, our method shows better generalization performance on novel relations, which shows that our proposed iterative joint training method effectively %narrows the semantic gap between the relations from different domains and 
        reduces the unwanted bias on source domain labeled data. (3) In addition, when a model has the tendency to cluster multiple relation into one, an unbalanced PR value (i.e., high \textit{rec.} and low \textit{prec.} in RSN-BERT) will be produced, which is undesired in real-world applications.
     
    \subsection{Incremental Learning of Novel Relations}
        \begin{figure}[t]
            \includegraphics[width=\columnwidth]{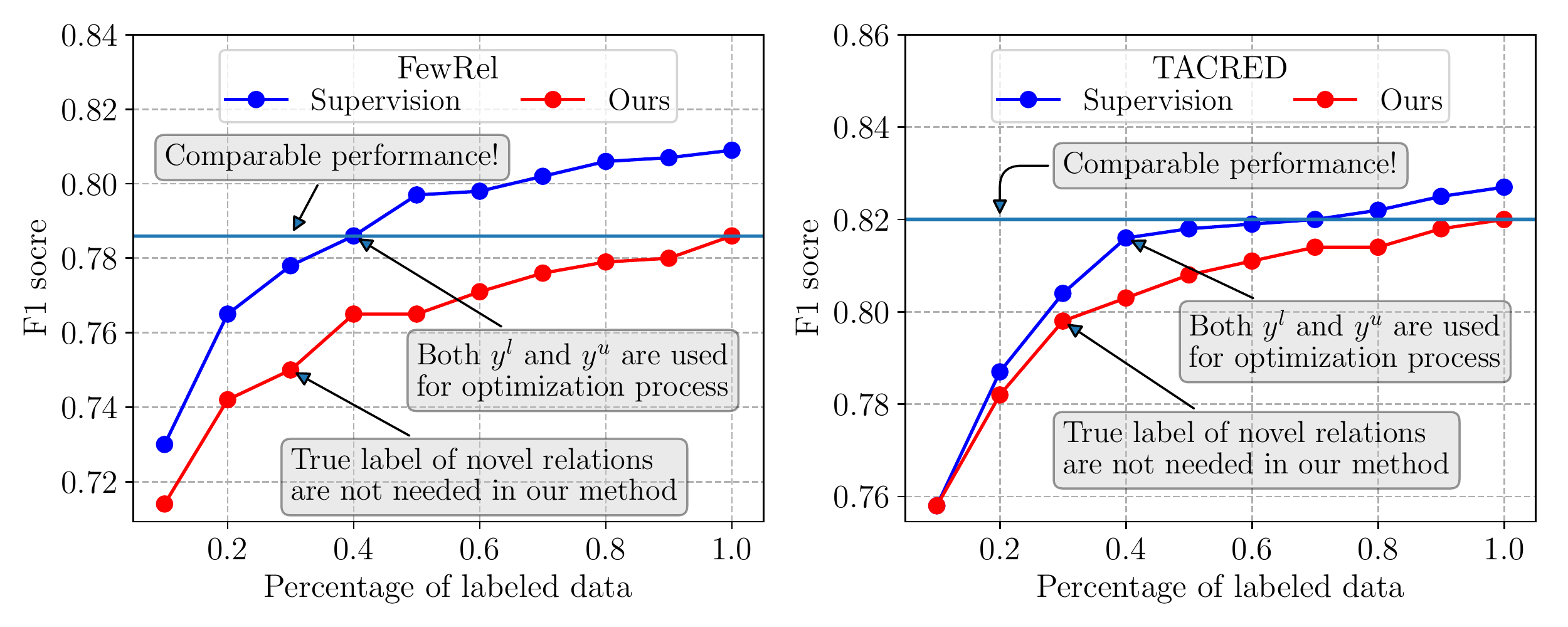}
            \caption{Model performance with different amounts of labeled data.}
            \label{fig:icr_learning}
        \end{figure}
        In this subsection, we evaluate the effectiveness of our incremental learning scheme and explore the influence of the amount of labeled data on model performance. We use BERT with a linear softmax classifer as the baseline for comparison. We train the baseline model using the labeled data of both pre-defined and novel relations, following the supervised learning paradigm. For our method, we still use only the labels of pre-defined relations. %Note that the model is required to discriminate both pre-defined and novel relations in this experiment.
        
        %From figure \ref{fig:icr_learning} we can observe the following: (1) The performance of the models improve gradually as labeled data increase. This is consistent with the general understanding that a large amount of labeled data can help improve the generalization of the model. (2) Our method can still maintain good performance when there is a lack of labeled data. This indicates that the proposed method is robust to the reduction of labeled data. (3) Our method achieves similar performance compared with the supervised baseline on  two experiments, which use 40\% labels of novel relations on FewRel dataset and 82\% on TACRED respectively. It indicates that we successfully achieve the incremental learning of novel relations. %, which often is desirable in real-world applications. 
        
        From figure \ref{fig:icr_learning} we can observe the following: (1) The performance of the models improve gradually as labeled data increase. Our method can still maintain good performance when there is a lack of labeled data. This indicates that the proposed method is robust to the reduction of labeled data. (2) Our method achieves similar performance compared with the supervised baseline on  two experiments, which use 40\% labels of novel relations on FewRel dataset and 82\% on TACRED respectively. It indicates that we successfully achieve the incremental learning of novel relations.

\section{Conclusions}
    In this work, we introduce a relation-oriented clustering method that extends the current unsupervised clustering-based OpenRE method. The proposed method leverages the labeled data of pre-defined relations to learn a relation-oriented representation from which the derived clusters explicitly align with relational classes. Iterative joint training method effectively reduces the unwanted bias on labeled data.
    %, improving both the supervised classification of the labeled data,  and the clustering of the unlabeled data.
    In addition, the proposed method can be easily extended to incremental learning of novel relations. Experimental results show that our method outperforms SOTA methods for OpenRE.
    
\section*{Acknowledgements}
The authors wish to thank the anonymous reviewers for their helpful comments. This work was partially funded by China National Key R\&D Program (No. 2018YFB1005104), National Natural Science Foundation of China (No.  62076069, 61976056), Shanghai Municipal Science and Technology Major Project (No.2021SHZDZX0103).

% Entries for the entire Anthology, followed by custom entries
\bibliography{anthology,custom}
\bibliographystyle{acl_natbib}
\appendix
\section{Hyperparameter Analysis}
        \label{app:Hyper}
        \begin{figure}[t]
            \includegraphics[width=\columnwidth]{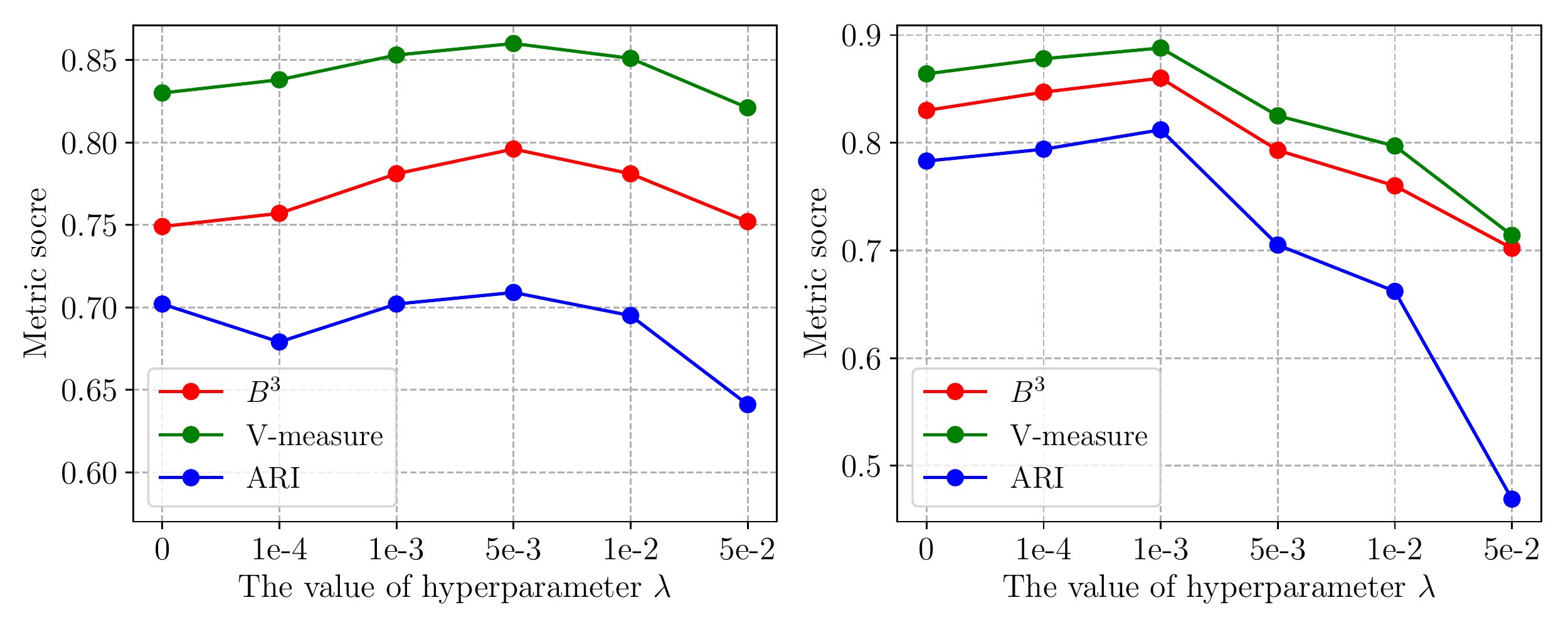}
            \caption{Model performance with different $\lambda$.}
            \label{fig:lambda}
        \end{figure}
        From the experimental results of ablation study, it can be seen that reconstruction loss and center loss have a great impact on the performance of the model. $\lambda$ is a key hyperparameter that balances the reconstruction loss versus center loss. In this section, we conduct experiments to study the influence of the value of $\lambda$ on the performance of the model. From Figure \ref{fig:lambda} we can see that: (1) When $\lambda$ gradually increases from 0, the center loss begins to affect the optimization. The model learns that instances with the same relation should be mapped to relatively close positions in the representation space, and the performance of the model gradually improves. (2) When the lambda exceeds a certain threshold, further increasing the $\lambda$ will leads to unwanted bias to the predefined relations, which will degrade the performance of the model.
        
    \section{Relation Representation Visualization}
    \label{app:visual}
    \begin{figure}[]
        \centering%使插图居中
        \includegraphics[width=\columnwidth]{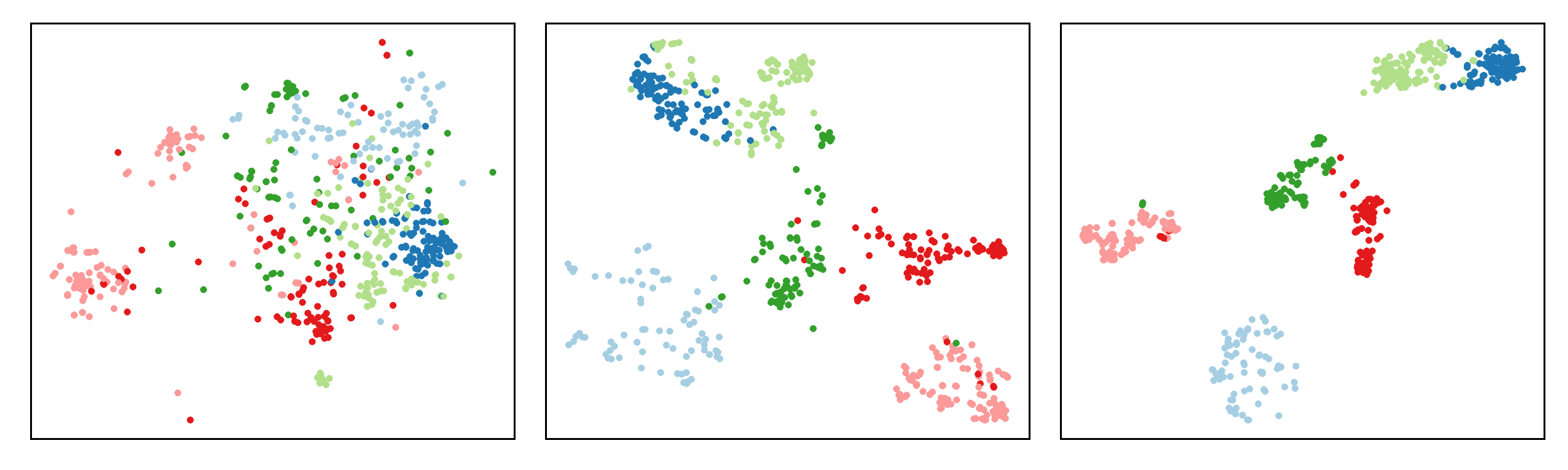}
        \caption{Visualization of the relation representation after t-SNE dimension reduction. The representations are colored with their ground-truth relation labels.
        These three sequentially illustrate the feature representation of initial sate, after reconstruction pre-training, and after training. All figures visualize the clustering result for 600 instances of randomly selected 6 novel relations on FewRel test dataset.}
        \label{fig:visual}
    \end{figure}   
    To intuitively show how the RoCORE method learns the constantly optimized relation-oriented representation, we visualize the relational representation with t-SNE \citep{JMLR:v9:vandermaaten08a}. The visualization results are shown in Figure \ref{fig:visual}. It is apparent that, before training (left), the relational representations are distributed randomly  at different locations in the semantic space. After pre-training (middle), the relational representations still are not tailored for the relations. For example, the instances with blue and light green colors may have similar syntactic or surface features and clustering them directly will lead to a poor result. After training (right), the relational representations are well separated and the 
   distribution is based on relation types. 
   \section{Detailed Results of Other Experiments}
   \label{app:other}
   In this section, the detailed results of ablation experiments and cross domain analysis are listed in Table \ref{tab:det_aba} and Table \ref{tab:det_cro} respectively.
        
        \begin{table*}
            \centering
            \resizebox{\linewidth}{!}{
            \begin{tabular}{ll lll lll l}
            \toprule
            \multirow{2}{*}{\textbf{Dataset}} & \multirow{2}{*}{\textbf{Method}} & \multicolumn{3}{c}{$B^3$} & \multicolumn{3}{c}{V-measure} & \multirow{2}{*}{ARI}\\
            \cline{3-8}
            & & Prec. & Rec. & $F_1$ & Hom. & Comp. & $F_1$\\
            \midrule
            \multirow{4}{*}{\textbf{FewRel}}
            &w/o center loss & $0.726_{32}$ & $0.774_{31}$ & $0.749_{31}$ & $0.818_{19}$ & $0.842_{18}$ & $0.830_{19}$ & $0.702_{38}$\\
            &w/o reconstruction & $0.512_{38}$ & $0.573_{17}$ & $0.540_{25}$ & $0.665_{24}$ & $0.689_{12}$ & $0.676_{16}$ & $0.495_{40}$\\
            &w/o CE & $0.662_{67}$ & $0.787_{54}$ & $0.719_{47}$ & $0.772_{41}$ & $0.844_{27}$ & $0.806_{28}$ & $0.617_{68}$\\
            &\textbf{RoCORE}& $0.752_{17}$ & $0.846_{09}$ & $\textbf{0.796}_{11}$ & $0.838_{10}$ & $0.883_{06}$ & $\textbf{0.860}_{07}$ & $\textbf{0.709}_{23}$ \\
            \hline \hline
            \multirow{4}{*}{\textbf{TACRED}}
            &w/o center loss & $0.818_{57}$ & $0.842_{37}$ & $0.830_{41}$ & $0.855_{41}$ & $0.867_{32}$ & $0.864_{32}$ & $0.783_{69}$\\
            &w/o reconstruction & $0.549_{35}$ & $0.483_{30}$ & $0.514_{31}$ & $0.589_{37}$ & $0.570_{28}$ & $0.579_{32}$ & $0.393_{54}$\\
            &w/o CE & $0.706_{45}$ & $0.776_{51}$ & $0.739_{47}$ & $0.753_{30}$ & $0.803_{37}$ & $0.777_{32}$ & $0.656_{85}$\\
            &\textbf{RoCORE}& $0.871_{42}$ & $0.849_{37}$ & $\textbf{0.860}_{35}$ & $0.895_{32}$ & $0.881_{20}$ & $\textbf{0.888}_{24}$ & $\textbf{0.812}_{64}$\\
            \bottomrule
            \end{tabular}}
            \caption{The detailed results of abalation study. The subscript represents the corresponding standard deviation (e.g., $0.749_{12}$ indicates $0.749\pm0.012$)}
            \label{tab:det_aba}
        \end{table*}    
        
\begin{table*}
            \centering
            \begin{tabular}{ll lll lll l}
            \toprule
            \multirow{2}{*}{\textbf{Task}} & \multirow{2}{*}{\textbf{Method}} & \multicolumn{3}{c}{$B^3$} & \multicolumn{3}{c}{V-measure} & \multirow{2}{*}{ARI}\\
            \cline{3-8}
            & & Prec. & Rec. & $F_1$ & Hom. & Comp. & $F_1$\\
            \midrule
            \multirow{3}{*}{$\textbf{F}\rightarrow \textbf{T}$}
            &RSN & 0.349 & 0.590 & 0.439 & 0.387 & 0.533 & 0.448 & 0.279\\
            &RSN-BERT & 0.337 & 0.866 & 0.486 & 0.400 & 0.777 & 0.528 & 0.352\\
            &\textbf{RoCORE} & $0.621_{28}$ & $0.602_{51}$ & $0.611_{34}$ & $0.642_{37}$ & $0.666_{32}$ & $0.654_{31}$ & $0.451_{65}$\\
            \hline \hline
            \multirow{3}{*}{$\textbf{F}\rightarrow \textbf{T}$}
            &RSN & 0.225 & 0.529 & 0.316 & 0.359 & 0.507 & 0.420 & 0.243 \\
            &RSN-BERT & 0.261 & 0.861 & 0.400 & 0.438 & 0.822 & 0.571 & 0.263\\
            &\textbf{RoCORE}& $0.687_{36}$ & $0.766_{46}$ & $\textbf{0.724}_{26}$ & $0.796_{22}$ & $0.836_{22}$ & $\textbf{0.815}_{16}$ & $\textbf{0.658}_{43}$\\
            \bottomrule
            \end{tabular}
            \caption{
            The detailed results of cross domain analysis. The subscript represents the corresponding standard deviation (e.g., $0.724_{26}$ indicates $0.724\pm0.026$)
            }
            \label{tab:det_cro}
        \end{table*}   
\end{document}